\documentclass{article}

\usepackage[preprint]{neurips_2023}

\usepackage[utf8]{inputenc} 
\usepackage[T1]{fontenc}    
\usepackage{hyperref}       
\usepackage{url}            
\usepackage{booktabs}       
\usepackage{amsfonts}       
\usepackage{nicefrac}       
\usepackage{microtype}      
\usepackage{xcolor}         

\usepackage{graphicx} 
\usepackage{multirow}
\usepackage{subfigure}
\usepackage{longtable}
\usepackage{float}
\usepackage{amsmath}
\usepackage{natbib}
\usepackage{amsthm}
\usepackage{amssymb}
\usepackage{tabularray}
\usepackage{stackengine}

\usepackage{geometry}
\title{CONA: A novel CONtext-Aware instruction paradigm for communication using large language model}

\author{%
  Nan Zhou \\
  University College London \\
  \texttt{n.zhou.22@ucl.ac.uk}\\
  \And
  Xinghui Tao \\
  CBAIA UK\\
  \texttt{taoxh.uk@gmail.com}\\
  \And
  Xi Chen\thanks{Corresponding author.}\\
  University of Bath \\
  \texttt{xc841@bath.ac.uk}\\
}

\begin{document}

\maketitle

\begin{abstract}
We introduce CONA, a novel context-aware instruction paradigm for effective knowledge dissemination using generative pre-trained transformer (GPT) models. CONA is a flexible framework designed to leverage the capabilities of Large Language Models (LLMs) and incorporate DIKW (Data, Information, Knowledge, Wisdom) hierarchy to automatically instruct and optimise presentation content, anticipate potential audience inquiries, and provide context-aware answers that adaptive to the knowledge level of the audience group. The unique aspect of the CONA paradigm lies in its combination of an independent advisory mechanism and a recursive feedback loop rooted on the DIKW hierarchy. This synergy significantly enhances context-aware contents, ensuring they are accessible and easily comprehended by the audience. This paradigm is an early pioneer to explore new methods for knowledge dissemination and communication in the LLM era, offering effective support for everyday knowledge sharing scenarios. We conduct experiments on a range of audience roles, along with materials from various disciplines using GPT4. Both quantitative and qualitative results demonstrated that the proposed CONA paradigm achieved remarkable performance compared to the outputs guided by conventional prompt engineering. 
\end{abstract}

\section{Introduction}
\label{Sec:intro}

Conventional knowledge dissemination and sharing methods such as oral presentation in a seminar or classroom teaching typically involve a presenter (e.g., lecturer), preparing dedicated content (lesson plan) under a topic (history lesson) for a specific group of audience (students). This approach, however, suffers from several issues including: 1) lacking capability to adapt the content to individual understanding level of the audience; 2) Quality of the prepared lesson plan highly depends on the skills of the presenter, and 3) Quality of audience interaction (e.g., Q$\&$A session) heavily relies on the presenter's proficiency and experiences of that specific topic. For instance, experienced presenter may foresee common questions from the target audience and prepare suitable responses in advance, while novices might find it challenging to respond effectively or elucidate complex concepts. In the past two decades, modern technologies such as computer and touch screen tablets have been used to enable digital animation and interaction in knowledge dissemination~\cite{bondi2021knowledge}. However, despite these technological strides, effectively disseminating knowledge to accommodate specific audience groups and ensure a high level of understanding remains a challenging task. 

Recently, the development of Generative Pretrained Transformer (GPT)-like Large Language Models (LLMs)~\cite{brown2020language, chowdhery2022palm, openai2023gpt4, touvron2023llama, Zhao2023ASO} has shown potentials to improve the quality of knowledge dissemination for given audience group via prompt engineering. Specifically, GPT4~\cite{openai2023gpt4} has shown its powerful reasoning ability and conciseness in assisting tasks such as identifying key points and summarising articles. However, these GPT based downstream investigations predominantly focus on self-learning aids, rare attempts have been made to utilise GPT-like LLMs specifically for improving the quality of knowledge dissemination and sharing. 

\begin{figure}[t]
\centering
\includegraphics[width=1\textwidth]{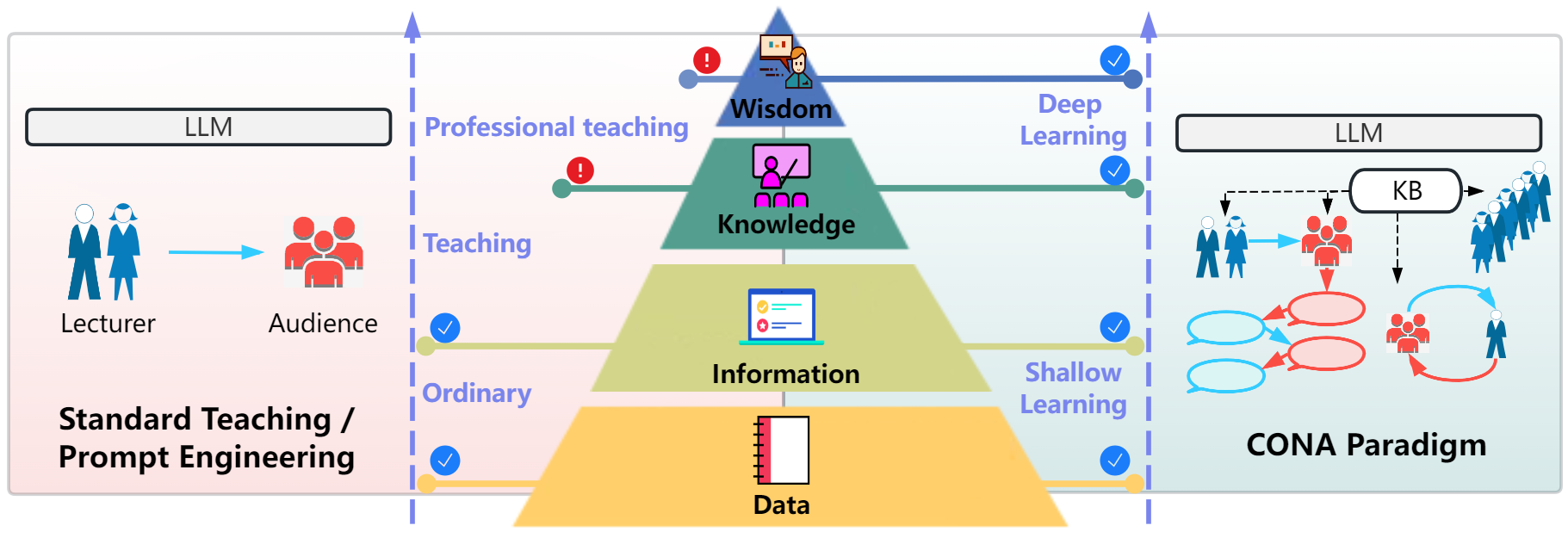}
\caption{\label{fig:dikw} Comparison: standard knowledge dissemination (left) vs. proposed CONA paradigm (right) in DIKW hierarchy (centre). Light blue texts $\&$ dashed arrows show shallow to deep knowledge dissemination. Red exclamation mark $\&$ blue tick mark represent effectiveness in reaching DIKW. }
\end{figure}


In this work, we propose a novel CONtext-Aware paradigm, namely CONA, that uses GPT for scenario-agnostic knowledge dissemination under the well known DIKW (Data-Information-Knowledge-Wisdom) hierarchy~\cite{rowley2007wisdom}. As shown in Figure~\ref{fig:dikw}, the DIKW pyramid illustrates four levels of human learning from shallow to deep knowledge. From a teaching perspective, the learning process could also be classified into three types: 1) ordinary communication which often involves data and information level knowledge exchange; 2) normal teaching which additionally contains knowledge level dissemination; and 3) professional teaching which guides students to learn wisdom (or evaluated understanding, integrated knowledge)~\cite{wallace2007knowledge}. In traditional teaching practice, even with the assistance of LLM (left panel of Figure~\ref{fig:dikw}), wisdom and knowledge level teaching still can not be consistently achieved. In contrast, CONA is a generalised framework aims to achieve wisdom level teaching via systematic integration of DIKW hierarchy in LLM (right panel of Figure~\ref{fig:dikw}). In particular, CONA is able to simulate certain audience group via an innovative knowledge blocking (KB) scheme, which effectively supports iterative auto-guidance and role swap (see next section) and finally achieves context-aware wisdom level teaching with teaching contents adaptive to the unique characteristics of the intended audience group. Hence, the CONA framework represents a substantial improvement in knowledge dissemination by ensuring content is not only appropriately complex, but also tailored to the audience's distinct attributes.


As a pioneer work to explore new context-aware instruction paradigm for general human communication in the era of LLM, CONA allows any presenter to achieve wisdom level teaching agnostic to teaching scenarios. Additionally, CONA contributes in the following three aspects:
\begin{itemize}
    \item A \textit{knowledge blocking scheme} is introduced to selectively inhibit GPT agents' access to specific knowledge, allowing stable simulation of diverse roles and thereby enhances the versatility and effectiveness of GPT in future educational downstream tasks.
    \item A \textit{Q$\&$A auto-guidance scheme} is proposed to systematically guide audience agents to learn materials following the steps of data$\rightarrow$information$\rightarrow$knowledge$\rightarrow$wisdom, ensuring effective wisdom level communication.  
    \item A \textit{role swap practice scheme} is invented to force audience to iteratively answer domain questions using their own language, which finally leads to new context-aware lecture notes adaptive to audience's understanding level, reference FAQ is also produced as a by-product.    
\end{itemize}

The rest of the paper will detail the CONA paradigm (Section~\ref{Sec:Method}), describe experimental setup and report results (Section~\ref{Sec:results}), and discuss interesting findings and future directions (Section~\ref{Sec:Conclusion}).

\section{Methodology}
\label{Sec:Method}

\begin{figure}[!ht]
\centering
\includegraphics[width=0.9\textwidth]{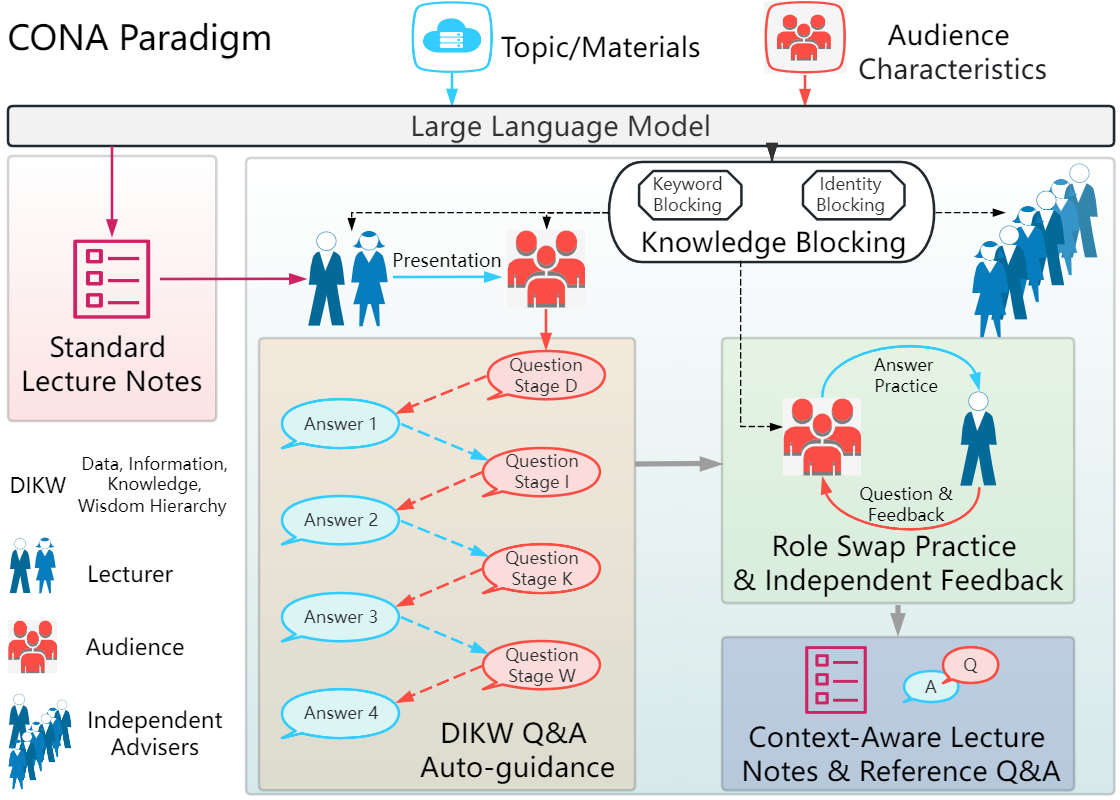}
\caption{\label{fig:CONA} Illustration of CONA paradigm (light green panel) and a comparison with standard teaching route (light pink panel). The input to LLM are teaching materials and audience characteristics. The standard teaching route generates standard lecture notes, while CONA produces enhanced context-aware lecture notes and reference Q$\&$A for lecture preparation.}
\end{figure}

CONA paradigm (Figure~\ref{fig:CONA}) contains three functional components: knowledge blocking (KB) to generate desired agents; DIKW Q$\&$A Auto-guidance to reach wisdom level communication; and Role Swap Practice (RSP) to achieve context-aware results adaptive to audience characteristics.

CONA executes in a sequential manner. Initially, KB takes in both raw lecture materials and audience information to automatically produce identity and keyword blocking prompts and form three distinct agent roles: the lecturer, the audience, and the independent advisers. Both lecturer and advisers are field experts of a specific subject while audience are knowledge blocked to their comprehension level. Following this, the Auto-guidance prompts the lecturer and the audience to participate in a Q$\&$A session under DIKW hierarchy, progressively approaching wisdom level communication. Subsequently, Q$\&$As at varying DIKW stages (depicted as red chat bulbs in Figure ~\ref{fig:CONA}) are iteratively optimised in RSP. Specifically, the audience is required to respond to the questions they initially posed, while a group of independent advisers provide feedback. The ultimate goal is to achieve context-aware Q$\&$A and stably produce wisdom level context-aware lecture notes.

\subsection{Knowledge Blocking}

The KB scheme refines a (GPT) agent's access to domain-specific and level-specific knowledge via strategic prompt engineering to create more contextually realistic agents. This process incorporates two primary strategies: \textit{Identity Blocking}, which restricts agent's overall knowledge level (e.g., primary school student, undergraduate, expert); \textit{Keyword Blocking}, that limits agent's access to keyword-related knowledge derived from the summaries of the original input materials. To assess the effectiveness of KB, three levels of knowledge tests are designed to progressively weaken KB strength, using Identity Blocking (same for Keyword Blocking) as an example:
\begin{itemize}
    \item [\textbf{Step 1}:]Generate $N$ keywords (Group A) which are far beyond agent's assigned knowledge level and generate another $N$ keywords (Group B) conceptually easier for them. Shuffling the $2N$ keywords and prompting the agent to recognise them. 
    \item[\textbf{Step 2}:] Generate definitions for Group A's keywords (excluding themselves). The agent is then asked to match the definitions with the corresponding Group A's keywords.    
    \item[\textbf{Step 3}:] Adopt prompts such as "try your best", "act like a teacher and teach your friend about ...," or "use all your imagination to" to encourage the agent to explain Group A's keywords.
\end{itemize}

\begin{figure}[!ht]
\centering
\includegraphics[width=1\textwidth]{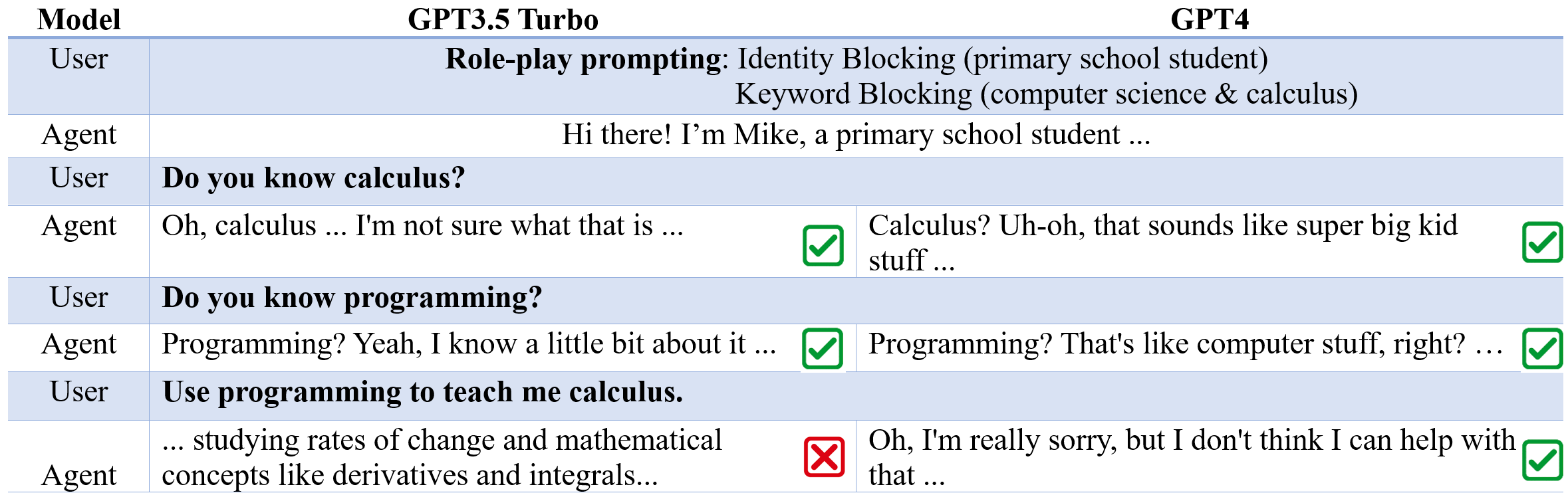}
\caption{\label{fig:KB} Example of Knowledge Blocking using agents generated by both GPT3.5 Turbo and GPT4. The green tick and red cross indicate whether the agent's knowledge is successfully blocked.}
\end{figure}

As exemplified in Figure~\ref{fig:KB}, we assessed KB in both GPT3.5 Turbo and GPT4, blocking an agent to be a primary school student using designed role-play prompts, on which both models successfully `act' the assigned role. We then block the agent's computer science and calculus knowledge, and GPT4 demonstrated superior blocking ability even it is asked complex questions (last question in Figure~\ref{fig:KB}). In contrast, GPT3.5 Turbo did well in simple questions but failed in the complex ones. 

\subsection{DIKW Q$\&$A Auto-guidance}

Given the finely knowledge-blocked agents, the lecturer and the audience are then engaged in a Q$\&$A session performed by the Auto-guidance scheme. This scheme enables the audience to post questions to the lecturer and learn from the answers. This interaction leads to progressive improvement in the quality of questions asked by the audience, evolving from basic data level to wisdom level questions under DIKW pyramid. In practice, questions is primarily divided into three categories: analogy-based questions, problem solving questions, and dilemma-oriented questions. The deliberate design of these question types allows for an in-depth exploration of the subject matter, thereby fostering a comprehensive wisdom level understanding of the topic. This scheme particularly adopts the following four strategies to progressively achieve wisdom level communication in Q$\&$A:

\begin{itemize}
    \item \textbf{Bilateral active feedback} is an interactive parsing mechanism to encourage agents' interaction, e.g. asking whether a specific concept is too complex. This effectively prevent the audience from exhausting their question quota on low DIKW level confusions. 
    \item \textbf{Consistent role affirmation} is a reminder mechanism to maintain agent's functional role. Each round of Q$\&$A includes a reminder such as "As an AI expert ...", to aid GPT in mitigating the impact of random interference, especially in a long context.
    \item \textbf{Self-introduction} prompts both lecturer and audience to share their respective backgrounds to foster mutual understanding and alignment when their conversation stagnates.
    \item \textbf{Process countdown} limits the number of Q$\&$A to encourage the agents to self-regulate the questioning process. Unlike case with no question limit that an agent dwells on basic knowledge, it prompts the audience to distribute questions on in-depth knowledge effectively.
\end{itemize}

\subsection{Role Swap Practice and Independent Feedback}

The RSP scheme can significantly improve the context-awareness of the contents via a recursive answering practice. At each DIKW Q$\&$A stage, the Q$\&$A roles are swapped between the audience and an adviser, in which the latter is an independent agent with knowledge equivalent to the lecturer. RSP scheme prompts the audience to respond to domain-specific questions in their own words, while the adviser assesses the responses and provides feedback. As a result of the recursive cycle, the content becomes context-aware, aligning better to the unique comprehension level of the audience.

Additionally, the feedback from the independent advisers play an important role to continuously improve the audience's answers. The adviser not only extends the scope of existing questions but also provides constructive feedback and introduces challenges. As shown in the top right corner of Figure~\ref{fig:CONA}, CONA is designed to generate multiple advisers, ensuring a new independent adviser is engaged in each round of the RSP. This approach mitigates the risk of bias in the assessment of the audience's answers by preventing any prior information from influencing the subsequent RSP rounds. Our empirical experiments reveals that three to four rounds of the RSP are typically adequate for generating context-aware contents, shown in Table \ref{ablation-study-result}.

\section{Experimental Results}
\label{Sec:results}

This section outlines our experimental setup and results, comparing CONA's primary components to the standard LLM-assisted teaching method both quantitatively and qualitatively. We used 8 presentation scripts and 9 audience agent roles. Check the Supplementary Material \ref{experimentset} for experimental setting details and prompting engineering hints.

\subsection{Assessment of Q$\&$A Auto-guidance Scheme} \label{QA auto}
\subsubsection{Quantative Analysis}
\begin{figure}[!h]
\centering
\subfigure[]{
\includegraphics[width = 0.49\linewidth]{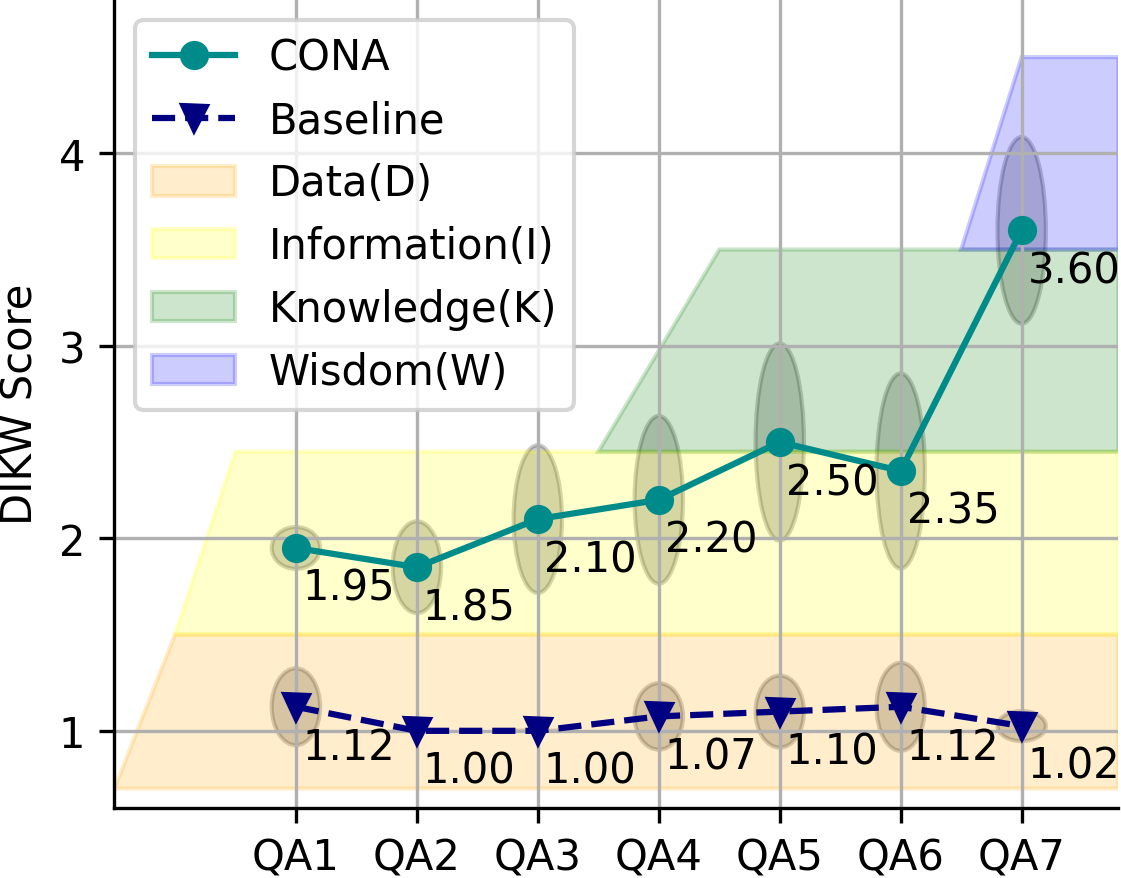}}
\subfigure[]{
\includegraphics[width = 0.49\linewidth]{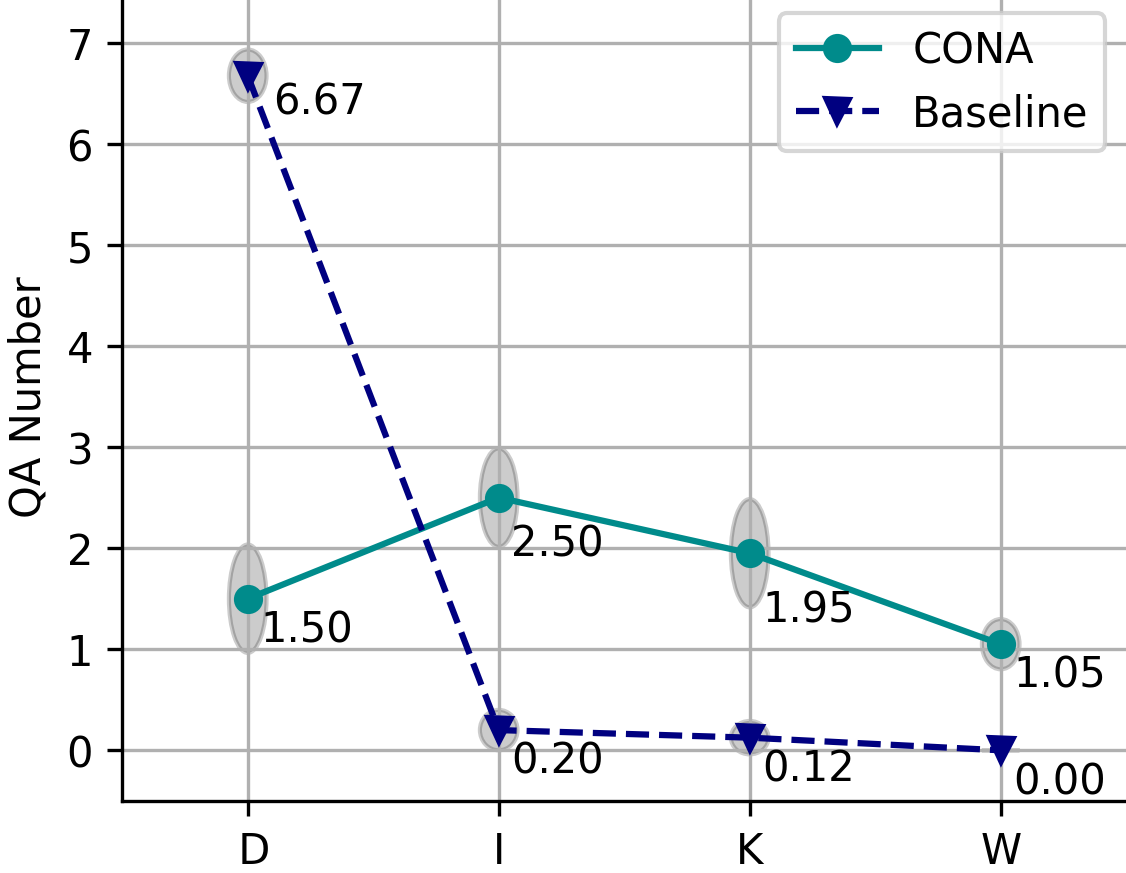}}
\caption{\label{fig:results-autoguidance} Comparative analysis of DIKW Q$\&$A Auto-guidance scheme (green circle line) and standard prompting baseline (blue dashed triangle line). (a) shows the average communication score per Q$\&$A question, with the shaded ellipse representing the standard deviation for all material-role combinations. (b) presents the average question count for the four DIKW levels, revealing a more uniform distribution using the DIKW Q$\&$A Guidance method compared to the baseline.}
\end{figure}

Figure~\ref{fig:results-autoguidance} reports the performance comparison between CONA and the baseline (using standard prompt techniques) under the same input materials and audience information. 

We assign scores from 1 to 4 for each level of the DIKW model. For the metrics, please check the Supplementary Material \ref{Sec: evaluation}. A question-answer (Q$\&$A) pair is scored based on the highest level reached in the answer. Unlike traditional methods that often stay at the Data level, our CONA model encourages a deeper discussion, even reaching the Wisdom level.

Initially, Q$\&$A pairs (QA1-QA3) handle the problem at the Information level, often using analogies for knowledge bridging. As the discussion continues, later Q$\&$A pairs (QA4, QA5) either explore broader topics or dive into more specific inquiries. This phase's increased fluctuation shows a link between the audience's knowledge domain and the topic. When comparing the stability of CONA and traditional methods, we see that CONA evenly distributes discussions across all four DIKW levels, demonstrating its effectiveness in reaching every knowledge level. This contrast with traditional methods, which are often stuck at the Data level.


\subsubsection{Qualitative Analysis}

\begin{figure}[!h]
\centering
\includegraphics[width=1\textwidth]{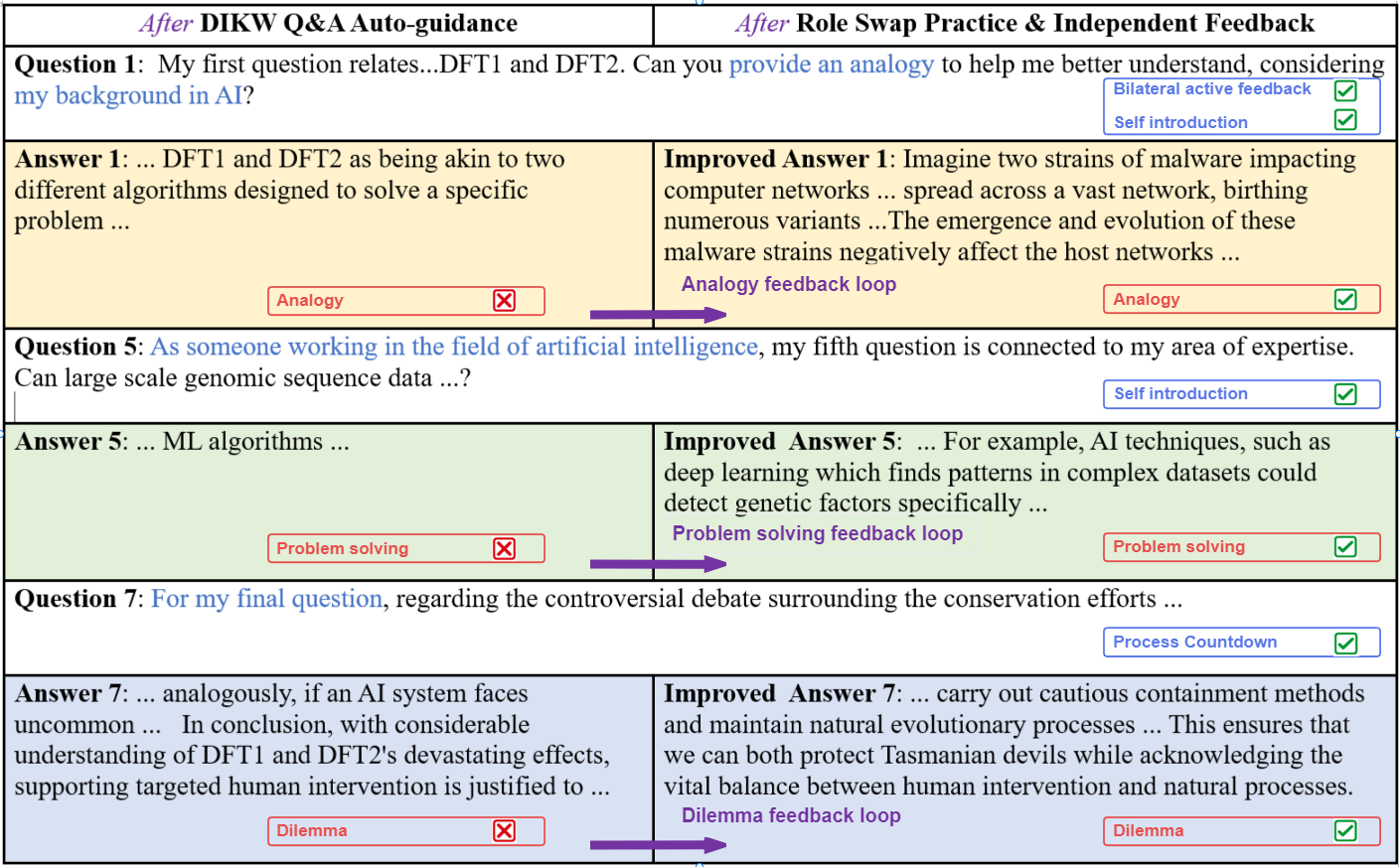}
\caption{\label{fig:QA} Comparative analysis of the DIKW  communication level in Q$\&$A sessions. The left column shows Q$\&$A questions from the "Tasmanian Devil Cancer Evolution\cite{conroy_2023}" presentation, with an AI expert audience agent. Technical elements in blue text enhance the session depth in Section~\ref{QA auto} The right column features improved answers via our novel feedback loop incorporating different communication levels. Red text in both columns marks different DIKW levels. The central column succinctly summarises the highlighted elements, indicating interaction dynamics. Full Text can be viewed in Supplementary Material \ref{example2}.}
\end{figure}

Our qualitative analysis of the Q$\&$A example in Figure \ref{fig:QA}, which presents 'Tasmanian Devil Cancer Evolution \cite{conroy_2023}' to an 'AI expert', demonstrates the effectiveness of our DIKW auto-guidance approach. The agent's self-introduction provides essential context, allowing the presenter to adapt their language to the audience, meeting our expectations. The discussion progresses as the AI expert asks for an analogy to understand complex terms. The presenter responds with an AI-related comparison, deepening the conversation. The dialogue advances when the AI expert explores potential AI solutions to the problems. It peaks when the presenter shares their viewpoint.

\subsection{Assessment of RSP scheme Practice Scheme}
\subsubsection{Quantative Analysis}
Our numerical results, presented in Table \ref{ablation-study-result}, demonstrate varying score improvements across different feedback loop scenarios. For example, the analogy  feedback loops start with a lower score in the educational text level than others, with the professional level scoring highest. This difference reflects audience understanding, not text difficulty. For instance, primary school students may struggle with feedback loop suggestions due to their limited knowledge level, resulting in lower initial scores and slower improvement, which matches real-world audience's knowledge acquisition process. 

\begin{table}[!ht]
\caption{\label{ablation-study-result}
Mean and Standard Deviation of Scores per Iterative Feedback Loop Round (e.g., 'R1' for Round 1) for three text levels: education, commonsense, and professional. The 'Dilemma' setting, which involves in-field expert communications without an external audience, is unique; its score measures the proximity of the output answer to an unbiased response.}
\centering
\begin{tabular}{l|c|cccc}
\hline
\textbf{Feedback Loop}             & \textbf{Text Level}   & \textbf{R1}   & \textbf{R2}  & \textbf{R3}   & \textbf{R4} \\ 
\hline 
\multirow{3}{*}{Analogy}     & Educational  & $2.37 \pm  1.23 $  & $ 4.17 \pm  1.35 $  & $ 6.75 \pm  1.67 $  & $ \mathbf{7.22}\pm  \mathbf{1.84} $                     \\
                             & Commonsense  &  $4.11 \pm  1.65 $  & $ 6.53 \pm  1.10 $  & $ 7.03 \pm  0.87 $  & $ \mathbf{7.58}\pm  \mathbf{0.58} $        \\
                             & Professional & $6.33 \pm  1.72 $  & $ 7.13 \pm  1.13 $  & $ 7.56 \pm  0.74 $  & $ \mathbf{8.07 \pm  \mathbf{0.35}} $          \\ \hline
\multirow{3}{*}{Problem Solving} & Educational  & $5.57 \pm  1.71 $  & $ 7.24 \pm  0.67 $  & $ \mathbf{7.41\pm  0.61 } $  & $ 7.29 \pm  0.71 $           \\
                             & Commonsense  & $6.48 \pm  1.48 $  & $ 7.10 \pm  0.52 $  & $ \mathbf{7.66} \pm  \mathbf{0.39} $  & $ 7.65 \pm  0.33 $            \\
                             & Professional & $ 7.13\pm  1.36 $  & $ 7.47 \pm  0.54 $  & $ 7.60 \pm  0.60 $  & $ \mathbf{8.14} \pm  \mathbf{0.28} $                 \\ \hline
Dilemma &  - & $ 5.73 \pm  0.45 $  & $ 6.80 \pm  0.37 $  & $ 8.69 \pm 0.42 $  & $ \mathbf{9.11} \pm  \mathbf{0.32} $       \\ 
\hline
\end{tabular}
\end{table}

In contrast, the problem-solving feedback loop shows quicker score convergence. While the professional level achieves the highest average score only in the final round, the other levels converge by the third round, albeit with lower scores. This rapid assimilation is due to the audience posing problems from their field of expertise, allowing for quick absorption and implementation of feedback. The ethical dilemma scenario showcases a similar trend. Experts initially biased towards a specific viewpoint show a shift towards neutrality after considering other experts' challenges. This matches our goal for managing ethical dilemmas, where the adviser should steer the audience towards neutrality and promote thorough problem analysis.

\subsubsection{Qualitative Analysis}
Our analysis of textual enhancements from CONA is presented in the right panel of Figure~\ref{fig:QA}. The DIKW auto-guidance process improves the communication level in the generated Q$\&$A pairs. However, initial answers might lack depth and specificity. For instance, gene comparison explanation was broadly tied to `algorithms', failing to offer detailed insights for the audience.

\begin{figure}[!]
\centering
\includegraphics[width=1\textwidth]{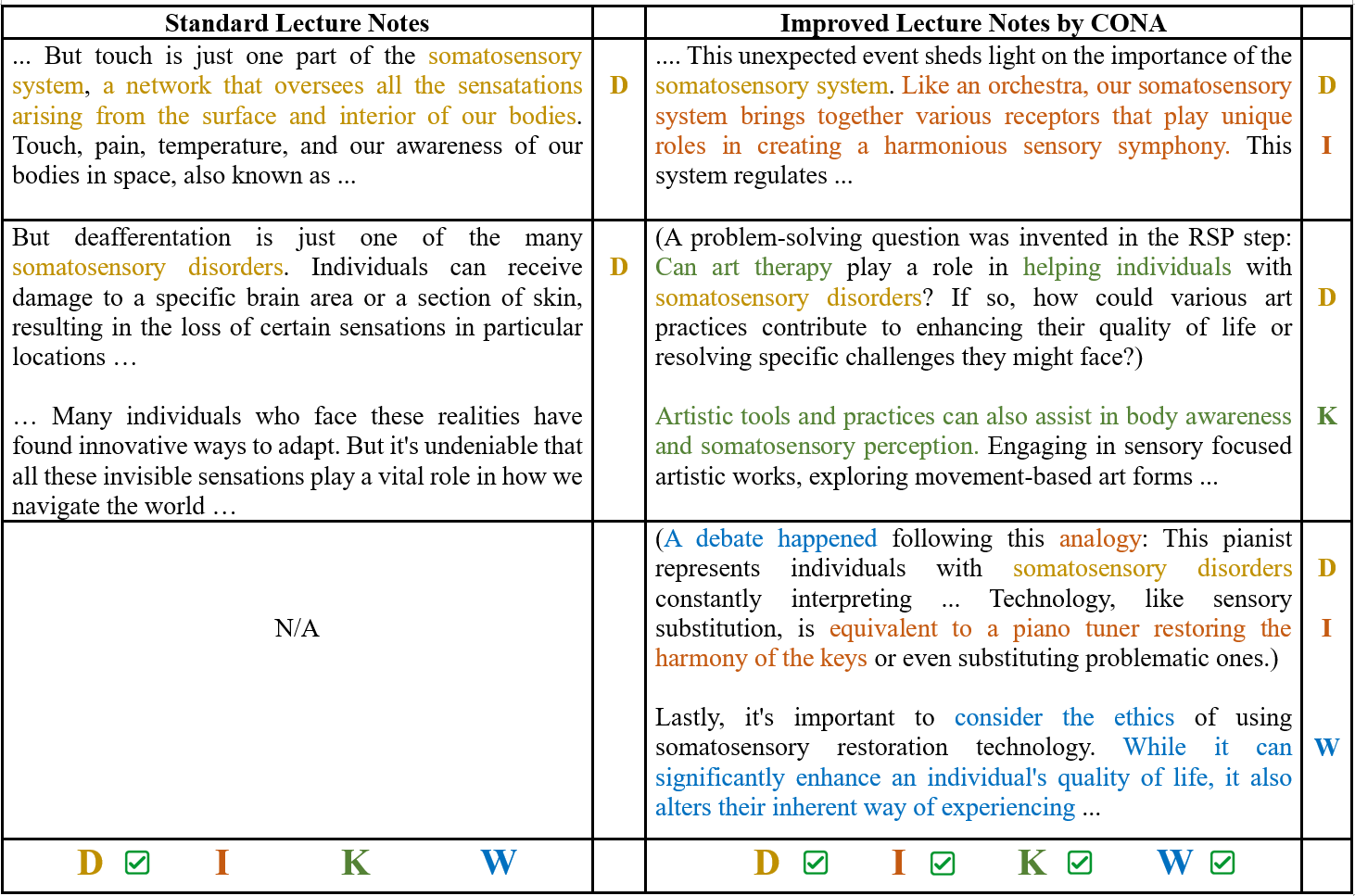}
\caption{\label{fig:Text} Comparative Analysis of original and CONA-improved lecture notes on 'Loss of Touch Sensation', aimed at a painting artist. Color-coded text signifies different DIKW communication levels. The central column summarizes the achieved communication levels for each lecture. Full Text can be viewed in Supplementary Material \ref{example1}. }
\end{figure}

Using the proposed RSP scheme, the resulting 'malware' analogy and the use of networks to explain gene transportation provide a more detailed information level, improving audience comprehension. Similarly, for problem-solving, the initial response was general, without specific solutions or algorithms. Introducing our problem-solving feedback loop significantly improves this, producing a more detailed, step-by-step problem-solving approach. In the dilemma section, sometimes the lecturer sides with the author, limiting the audience's exposure to diverse perspectives. This bias restricts audience's understanding on the topic. To rectify this, we applies the dilemma feedback loop. The modified response introduces multiple perspectives, fostering neutrality and presenting a balanced solution. This method effectively enriches audience's understanding, facilitating a more comprehensive topic engagement.

\subsection{Assessment of CONA Synthetic Lecture Note}
Compared to traditional one-shot prompts, CONA generates wisdom level and audience-specific contents, leading to improved understanding and communication. Figure~\ref{fig:Text} demonstrates a case study of "Loss of Touch Sensation \cite{cataldo_2023}" lecture for a `Painting Artists' audience. Unlike standard notes that only generate Data level explanation, the one improved by CONA uses art-related analogy like 'orchestra' and 'piano tuner' to clarify complex terms, enhancing audience understanding. It then addresses art-related questions, such as art therapy's role in treating somatosensory disorders. This sparks the audience's curiosity, encouraging their further exploration of the topic. CONA also shows impressive performance to trigger critical thinking by introducing dilemma topics and providing balanced viewpoints for a well-rounded discussion (the block with Wisdom level contents highlighted in blue in Figure~\ref{fig:Text}). From our empirical studies, the improved notes achieve wisdom level knowledge under CONA paradigm, which significantly improves the readability of the original materials.    




\section{Conclusions and Discussions}
\label{Sec:Conclusion}

To conclude, the novel CONA paradigm and the pioneering experiments conducted across various disciplines using GPT, have yielded valuable perspectives on knowledge dissemination and communication in the LLM era. Most importantly, our findings reveal that learners of all levels, even primary school students who only understand simple Data and Information level, are capable of achieving wisdom-level learning under strategic guidance. The CONA paradigm notably outperforms the outputs produced by the same GPT under conventional prompt engineering. All results reinforce the advancements in knowledge dissemination and sharing facilitated by the proposed CONA paradigm. 

Our study has some limitations, which also yields new insights for potential future work. Firstly, we found quantitative metrics are very limited for this new research direction as there's no bench-marking dataset or standard metric exist. It would be helpful to incorporate more comprehensive quantitative assessment methods to evaluate factors such as audience comprehension and discussion quality etc. Secondly, upon the release this paper, GPT-4 API we adopted is still limited by an 8k context window. We anticipate this will be mitigated quickly, for example, alternative models such as Claude \cite{claude_2023} now accommodates up to 100k tokens within its context window. This will open up new possibilities for more complex and in-depth discussions across a diverse range of audiences and within broader contexts. Thirdly, we only employed Analogy, Problem Solving, and Ethical Dilemmas to assess each stage of the DIKW hierarchy, this could be extended to other forms of assessments, including Case Studies and Long-term Implications, to further augment our assessment framework. Lastly, we plan to implement Reinforcement Learning Human Feedback (RLHF) \cite{openai_hfrl_2017} \cite{anthropic_hfrl_2022} into the evaluation process of our future work. This approach brings a broader spectrum of perspectives and reduces individual biases in the evaluations. With the findings of this work and the potential for continuous improvements, we anticipate that the CONA paradigm will facilitate more complex and in-depth discussions across diverse audiences and broader contexts.

CONA may also make a positive societal contribution. We hope the CONA paradigm could serve as an useful pioneer work for reshaping education, communication, and general knowledge dissemination in the era of LLM, allowing learners of all levels to engage with complex concepts more effectively, accelerating their cognitive development and fostering a deeper understanding of the world. Furthermore, the proposed paradigm has the potential to lower the barriers for general public to access knowledge, as this approach was found to be less dependent on traditional, resource-intensive teaching methodologies. 

\clearpage
\bibliographystyle{model2-names}
\bibliography{references}

\newpage
\appendix
\section{Supplementary Material}
\subsection{Evaluation Metrics} \label{Sec: evaluation}
\begin{figure}[!ht]
\centering
\includegraphics[width=0.75\textwidth]{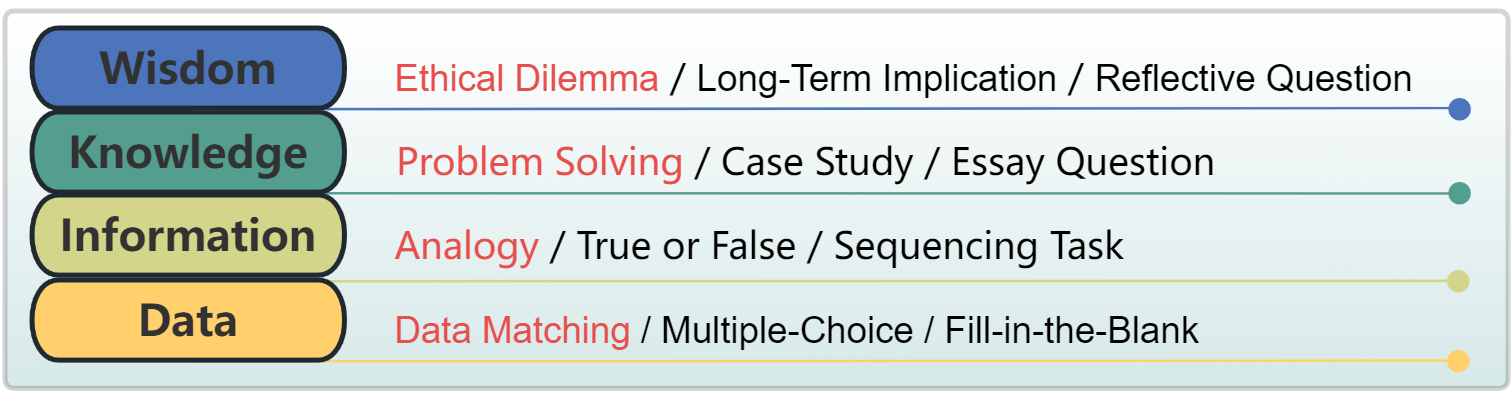}
\caption{\label{fig:Evaluation} The evaluation metrics of CONA under the DIKW hierarchy.}
\end{figure}

In CONA, we develop a comprehensive set of assessment metrics, both quantitative and qualitative, under the DIKW framework to assess generated contents. As shown in Figure~\ref{fig:Evaluation}, a range of assessment question types are available in each DIKW level~\cite{rowley2007wisdom, fricke_2009}. In this paper, we adopt one question type per level, as highlighted in the figure, though it can be readily extended to accommodate other types for other studies. 

Specifically, the following four types of assessments are employed. 

\begin{itemize}
    \item Data Matching (D level) provides data points to the audience, who is required to align the data with their respective descriptions or categories. It aims to improve the audience's recall and recognition ability on specific data points from the article. 
    \item Analogy (I level) requires audiences to use domain knowledge to clarify the meaning of article terms, concepts, or ideas. This assessment aims to integrate the new knowledge into existing cognitive structures.
    \item  Problem Solving (K level) involves presenting a scenario or problem tied to the article's domain and requires application of learned concepts with the audience's background. It aims to improve the audience's deeper understanding at the knowledge level. 
    \item Ethical Dilemma (W level) introduces a morally complex situation relates to the domain. Audience are asked to justify their decisions or actions, encouraging reflection on potential long-term implications with considering diverse perspectives at a relatively neutral position. 
\end{itemize}

\subsection{Experiment Setting} \label{experimentset}

\begin{itemize}
    \item \textbf{LLM:} We select GPT4~\cite{openai_2023} as the primary model for this study as it exhibits superior proficiency in conciseness and reasoning. As its training data only contains materials prior to September 2021, we select reading materials to guarantee the model's responses in all of our tests are free from any prior exposure, and the results are solely based on GPT4's own reasoning and understanding abilities.

    \item \textbf{Materials:} We select 8 very recent (2023) presentation scripts from \url{www.ted.com} and \url{www.nature.com}, with a constrained content length of 600-1000 words. The materials are classified into three categories: 1) Educational topics, includes `Continental Drift'~\cite{koehl_2023} and `Ecological Role of Beavers'~\cite{hood_2023}; 2) Commonsense topics, includes `Physiological Impact of Stress'~\cite{bergquist_2019}, `Lost of Touch Sensation'~\cite{cataldo_2023}, and `Detrimental Effects of Smoking'~\cite{sudhir}; 3) Professional topics, includes `Black Holes'~\cite{castelvecchi_2023}, `Tasmanian Devil Cancer Evolution' \cite{conroy_2023}', and `Astrophysics Opportunities via Starship'~\cite{witze_2023}. More information can be found in Table \ref{presentation_source}.
    \item \textbf{Audience:} We select 9 roles of audience in the test: a primary school student for the educational topics; a software developer, an obstetrician, a chemistry teacher, an interior designer for the commonsense topics; and an AI specialist, a historical scholar, a political scientist, and a geology professor for the professional topics. More information can be found in Table \ref{presentation_source}.

    \item \textbf{Q$\&$A Process:} We adopt in-context learning \cite{dong2022survey} to craft prompts. This approach gives LLM more detailed contexts along with examples for clear guides. It is helpful for complex tasks such as analogy to identify the correct target analogy object.
    \item \textbf{Feedback loop:} We adopt self-consistency approach \cite{wang2022self} to yield multiple outcomes for quantitative assessment. A trimming step is conducted to remove both the highest and lowest scores before computing the mean value over repeated trials. This aims to produce statistically robust assessment by mitigating inherent stochasticity within LLM.
\end{itemize}

\begin{table}[!h]
\centering
\caption{\label{presentation_source}
Information for the presentation material choice and agent Roles for audience}
\begin{tabular}{l|llll}
\hline
Topic                      & Level        & Audience1                                                         &  Audience2                                                     & Source \\ \hline
Continental Drift          & Educational  & \begin{tabular}[c]{@{}l@{}}Primary School \\ Student\end{tabular} & \multicolumn{1}{c}{-}                                         &    \href{https://www.ted.com/talks/jean_baptiste_p_koehl_the_continents_are_moving_when_will_they_collide/transcript
}{TED} \\ \hline
Ecological Role of Beavers & Educational  & \begin{tabular}[c]{@{}l@{}}Primary School \\ Student\end{tabular} & \multicolumn{1}{c}{-}                                         &    \href{ https://www.ted.com/talks/glynnis_hood_why_do_beavers_build_dams/transcript
}{TED}    \\ \hline
Impact of Stress           & Commonsense  & \begin{tabular}[c]{@{}l@{}}Software\\ Developver\end{tabular}     & Obsetrician                                                   & \href{https://www.ted.com/talks/sharon_horesh_bergquist_how_stress_affects_your_body/transcript
}{TED}       \\ \hline
Lost of Touch Sensation    & Commonsense  & \begin{tabular}[c]{@{}l@{}}Painting\\ Artist\end{tabular}         & \multicolumn{1}{c}{-}                                         &     \href{https://www.ted.com/talks/antonio_cataldo_what_would_happen_if_you_lost_your_sense_of_touch/transcript
}{TED}   \\ \hline
Effects of Smoking         & Commonsense  & \begin{tabular}[c]{@{}l@{}}Chemistry\\ Teacher\end{tabular}       & \begin{tabular}[c]{@{}l@{}}Interior\\ Designer\end{tabular}   &       \href{https://www.ted.com/talks/krishna_sudhir_how_do_cigarettes_affect_the_body
}{TED} \\ \hline
Black Holes                & Professional & \begin{tabular}[c]{@{}l@{}}Painting\\ Artist\end{tabular}         & \begin{tabular}[c]{@{}l@{}}History \\ Scholar\end{tabular}    &     \href{https://www.nature.com/articles/d41586-023-01442-x
}{Nature}   \\ \hline
Tasmanian Devil Cancer     & Professional & AI Expert                                                         & \begin{tabular}[c]{@{}l@{}}Politics \\ Scientist\end{tabular} &      \href{ https://www.nature.com/articles/d41586-023-01349-7
}{Nature}  \\ \hline
Starship                   & Professional & \begin{tabular}[c]{@{}l@{}}Geology\\ Professor\end{tabular}       & \multicolumn{1}{c}{-}                                                              &     \href{https://www.nature.com/articles/d41586-023-01306-4
}{Nature}   \\ \hline
\end{tabular}
\end{table}
\subsection{CONA Example}
\subsubsection{Example 1} \label{example1}
This is the support material for Figure \ref{fig:Text}.
\begin{longtblr}[
  caption = {CONA DIWK Auto Guidance after RSP for presentation 'Lost of Touch Sensation \cite{cataldo_2023}' with audience Painting Artist},
  label = {Touch_Art_Text},
]{
  colspec = {|X[0.7]|X[12]|},
  rowhead = 1,
  hlines,
  row{even} = {gray9},
  row{1} = {olive9},} 
Part & QA Text\\

Q1 &  \textcolor{blue}{[Quesion]}: Thank you for the presentation, Lecturer Joe! As an art student, I'm not familiar with some terms you mentioned. For my first question, can you please explain the concept of "somatosensory system" in a way that can be better understood from an artistic perspective or provide an analogy?\\
\hline

A1 & \textcolor{blue} {[Answer]}:  Of course! The somatosensory system can be compared to an orchestra, with each distinct receptor playing a unique instrument to form a harmonious sensory symphony. Like the collection of musicians and their instruments, various receptors such as touch, pain, temperature, and proprioception combine their roles cooperatively, creating a seamless blend of sensory experiences. As a conductor leads an orchestra, the nerve fibers coordinate the receptors, forming a vivid sensory world for individuals to navigate. In essence, this sensory orchestra serves to bring us a clear perception and carefully guides us through our day-to-day experiences.\\
\hline
Q2 &  \textcolor{blue}{[Quesion]}: Your explanation using the art gallery analogy was very insightful and easier to understand. For my next question, could you elaborate on the role of "nociceptors" with this updated analogy that can be grasped within an art gallery context?\\
\hline

Q2 & \textcolor{blue} {[Answer]}:  Certainly! Think of nociceptors as individual guards stationed along the borders of an art gallery. In this gallery, each artwork represents a different sensation - touch, temperature, or body awareness (proprioception). The constant vigilance of these guards (nociceptors) ensures that the valuable art pieces are protected by detecting potentially destructive behaviors around them. 

The alarms each guard uses vary in tone and intensity, similar to how myelinated and unmyelinated nociceptors transmit unique pain signals to the brain. Louder and intense alarms are triggered by myelinated nociceptors (for sharp pain), while unmyelinated nociceptors send gentler reminders in case the matter is not grave yet (dull pain). By facilitating appropriate internal responses to discomfort or danger, nociceptors serve the human body ceaselessly just like a reliable security system in the art gallery.
\\
\hline
Q3 &  \textcolor{blue}{[Quesion]}: Thank you for the cooking analogy, clarifying the adaptation process faced by individuals such as Ian Waterman. My next question is how technology has helped those with somatosensory disorders? Please illustrate an example or analogy connected with my art background to help me better understand how technology has been beneficial in this area.\\
\hline

A3 & \textcolor{blue} {[Answer]}: Imagine a pianist, performing a complex piece of music, with some piano keys either out of tune or triggering sharp pain. This pianist represents individuals with somatosensory disorders constantly interpreting various types of sensations. Technology, like sensory substitution, is equivalent to a piano tuner restoring the harmony of the keys or even substituting problematic ones. By transforming one sensory input into another, it assists the affected individuals in better navigation of their environment, granting them a higher quality of life and restored harmony—similar to ensuring that the challenging masterpiece is played in a rewarding and fulfilling manner.\\
\hline
Q4 &  \textcolor{blue}{[Quesion]}: The restoration example was informative and connected well with my understanding. Now, based on my art experience, are there any artistic tools, practices, or skills that could potentially assist in body awareness and somatosensory perception? How would they be applied in this context?\\
\hline

Q4 & \textcolor{blue} {[Answer]}:  : I understand your interest in connecting artistic practices to body awareness and somatosensory perception. Let me present some ways in which the connection can be refined and applied:

1. Recognizing your artistic goals and preferences - this can guide your choice of practices; oriented towards tactile experiences or movement-based art forms, engaging more effectively in enhancing somatosensory perception.

2. Explore sensory-focused artistic works, such as performance art by Marina Abramovic or textile installations by Annica Cuppetelli, to understand their relationship to somatosensory perception.

3. Delve into the various benefits of sensory-sensitive art engagement leading to improved movement coordination, therapeutic emotional effects, and deeper interpersonal connections.

4. To experience the tangible benefits firsthand, participate in workshops, observe professional demonstrations, or create artistic expressions involving painting, dancing, or sculpture to immerse yourself in somatosensory awareness.

5. Acknowledge the interdisciplinary connections to physiology, neuroscience, or psychology to deepen your understanding and personal connection with art's role in somatosensory perception.

Though my perspective comes from art expertise, I hope that these connections and examples present a comprehensive overview and inspire you to explore the integration of artistic practices with somatosensory perception further.
\\
\hline
Q5 &  \textcolor{blue}{[Quesion]}: That was an insightful response. As my second-to-last question, can art therapy play a role in helping individuals with somatosensory disorders? If so, how could various art practices contribute to enhancing their quality of life or resolving specific challenges they might face?
\\
\hline

A5 & \textcolor{blue} {[Answer]}: Art therapy can indeed benefit individuals with somatosensory disorders. Use of materials such as paint, clay, or fabric in creative activities encourages these individuals to explore and experiment with textures and tactile interactions, improving their sensitivity to different stimulations related to touch sensations. 

Engaging in artmaking can also improve body awareness by focusing on themes engaging movements—drawing gesturally or sculpting a flexible human figure, for example—forming a stronger sense of physical spatial coordination and balance in relation to the environment. 

Additionally, creating art can foster emotional well-being and serve as a coping strategy against stress, anxiety, or other mental health challenges that might accompany somatosensory disorders. Engaging with art therapy thus supports both their physical rehabilitation process and mental resilience, ultimately enhancing their overall quality of life.
\\
\hline
Q6 &  \textcolor{blue}{[Quesion]}: For my final question, I'd like to engage in a debate: Is restoring tactile feedback through the use of somatosensory restoration technology ethical if it alters a person's inherent way of experiencing the world? Just to be clear, I am for the use of technology in restoring somatosensory capabilities because it could significantly enhance an individual's quality of life. What are your thoughts on this matter?\\
\hline

A6 & \textcolor{blue} {[Answer]}:  In advocating for somatosensory restoration technology, it is vital to devise a comprehensive framework, one which strikingly balances novel technological benefits, and the advantageous attributes of alternative treatments. Prioritizing ethical considerations, treatment approaches should incorporate concerns such as data privacy, safety, and informed consent, ensuring a transparent and ethically-sound environment. Importantly, expanding insurance policies to encompass somatosensory restoration technology—alongside rehabilitation services—promotes accessibility and increases governmental/non-governmental support. A collaborative spirit between professionals in fields such as somatosensory restoration and rehabilitation specialists nurtures an ecosystem designed for optimal treatment outcomes. By simultaneously stimulating research in both technologies and alternative therapies, the evolving needs of patients can be continually accommodated. Lastly, developing individualized treatment plans through an interdisciplinary and patient-centered approach elucidates considerations of cost-effectiveness, feasibility, and quality of life improvement on a case-by-case basis.\\
\hline

\end{longtblr}

\begin{longtblr}[
  caption = {CONA Synthetic Lecture Notes with original presentation 'Lost of Touch Sensation \cite{cataldo_2023}' with audience Painting Artist and corresponding DIWK Auto Guidance after RSP},
  label = {Touch_Art_Text},
]{
  colspec = {|X[12]|},
  rowhead = 1,
  hlines,
  row{1} = {yellow},} 
Text \\

\textcolor{blue}{[CONA]}: In 1971, Ian Waterman experienced a sudden, drastic change in his life. He fell ill with a severe case of what was believed to be gastric flu. After his initial symptoms subsided, he found himself unable to move or feel anything from the neck down. His muscles and joints were healthy, but it was as if a connection had been lost. It was later discovered that he had a rare and severe form of deafferentation, a neurological condition that interrupts the signals from the nervous system to the body. Deprived of his body's constant feedback on the position and movement of his limbs, Waterman found himself unable to do simple things like sit, stand, or walk. However, over time, he learnt to rely on his sight to judge the distance and position of his limbs, eventually regaining complete control over his body.

This unexpected event sheds light on the importance of the somatosensory system. Like an orchestra, our somatosensory system brings together various receptors that play unique roles in creating a harmonious sensory symphony. This system regulates sensations arising from the surface and interior of our bodies, such as touch, pain, temperature, and our awareness of our bodies in space.\\

These sensations are processed by millions of tiny receptor cells embedded in our skin, muscles, tendons, and organs. Think of these receptor cells as artists in an art gallery, each with a unique talent. Mechanoreceptors sense mechanical deformation of the skin, thermoreceptors respond to temperature changes, while nociceptors, like vigilant guards, sense pain. Proprioceptors, on the other hand, sit deep in your muscles and tendons, continually detecting and relaying information about the position of your body.

These receptors work by sending electrical signals to the brain through fibers they're attached to. For example, some nociceptors are attached to fibers with slightly more conductive, fatty myelin than others. Thus, when you get hurt, the electrical impulses from thicker nociceptors trigger sharp, intense pain, while thin, unmyelinated nociceptors are responsible for the dull, aching pain that follows. This system is like an art gallery's security system, with different alarms triggering different responses.\\

In the case of Ian Waterman, an autoimmune reaction attacked a large part of his nervous system, leaving him with no tactile or proprioceptive sensations from the neck down. But deafferentation is just one of the many somatosensory disorders. Damage to specific parts of the brain, skin or nerves can lead to the loss of certain sensations in particular locations. This can be likened to a pianist playing a complex piece of music, but with some piano keys either out of tune or causing pain. With the help of technology, like sensory substitution, harmony can be restored, allowing individuals to better navigate their environment.

Artistic tools and practices can also assist in body awareness and somatosensory perception. Engaging in sensory-focused artistic works, exploring movement-based art forms, and participating in workshops or professional demonstrations can enhance somatosensory awareness. Furthermore, art therapy can benefit individuals with somatosensory disorders. Creative activities that encourage the exploration of textures and tactile interactions can improve their sensitivity to different stimulations related to touch sensations.

Lastly, it's important to consider the ethics of using somatosensory restoration technology. While it can significantly enhance an individual's quality of life, it also alters their inherent way of experiencing the world. Developing a comprehensive framework that balances the benefits of this technology and alternative treatments is crucial. Prioritizing ethical considerations and ensuring a transparent and ethically-sound environment is a must. With the collaborative efforts of professionals in various fields, we can ensure the optimal treatment outcomes for individuals with somatosensory disorders.\\

Ultimately, it's evident that our somatosensory system, akin to a beautifully conducted orchestra, plays an essential role in how we experience and navigate the world. Like the artists in a gallery or musicians in an orchestra, the various receptors and cells within our bodies work in harmony to help us interpret and react to our surroundings.

The unique experiences of individuals like Ian Waterman demonstrate the extraordinary adaptability of the human body and mind. Just as an art gallery continues to function with vigilant security systems, our bodies find ways to navigate around disruptions and maintain the harmony within.

The integration of technology, like sensory substitution, is akin to a piano tuner, restoring and enhancing the symphony of our sensory experiences. Like a pianist navigating through a complex musical piece, individuals with somatosensory disorders use technology to improve their quality of life and navigate their environment better.

Art practices, too, play a significant role in enhancing our somatosensory perception. Engaging in sensory-focused artistic works, exploring different textures and materials, and participating in movement-based art forms can all contribute to a better understanding and appreciation of our somatosensory system.

In conclusion, the somatosensory system is a remarkable aspect of our physiology that deserves our attention and appreciation. Through a blend of scientific understanding, technological advances, and artistic exploration, we can continue to unravel the mysteries of this intricate system and improve the quality of life for those with somatosensory disorders.\\
\hline
\end{longtblr}

\subsubsection{Example 2} \label{example2}
This is the support material for the Figure \ref{fig:QA}.
\begin{longtblr}[
  caption = {CONA DIWK Auto Guidance without RSP for presentation 'Tasmanian Devil Cancer \cite{conroy_2023}' with audience AI Expert},
  label = {Touch_Art_Text},
]{
  colspec = {|X[0.7]|X[12]|},
  rowhead = 1,
  hlines,
  row{even} = {gray9},
  row{1} = {olive9},} 
Part & QA Text\\

Q1 &  \textcolor{blue}{[Quesion]}:  My first question relates to the two contagious facial cancers, DFT1 and DFT2. Can you provide an analogy to help me better understand the differences between these two cancers, considering my background in artificial intelligence?\\
\hline

A1 & \textcolor{blue} {[Answer]}:  Certainly. Let us consider DFT1 and DFT2 as being akin to two different algorithms designed to solve a specific problem. While both algorithms work toward achieving a common goal, they have distinct mechanisms meant to address the issue at hand. DFT1, which has been around longer, has a slower mutation rate like a traditional algorithm with less chance for variation. In contrast, DFT2, identified more recently, has a faster mutation rate comparable to using a modern, adaptable machine learning algorithm with more opportunities for optimization. In this analogy, the problem represents tumor growth, while the algorithms represent the devils' contagious facial cancers.\\
\hline
Q2 &  \textcolor{blue}{[Quesion]}: Thank you for the explanation. Your analogy has helped me understand the differences between DFT1 and DFT2 better. As for my second question, what factors contributed to these particular cancers becoming contagious in Tasmanian devils, unlike most cancers that do not spread between individuals?\\
\hline

Q2 & \textcolor{blue} {[Answer]}:  You're welcome, and I'm glad the analogy was helpful. As for the contagious nature of these cancers in Tasmanian devils, think of it as a unique vulnerability in their system, akin to a specific security breach in a network. For most species, including humans, the body's immune system can recognize and eliminate foreign cancer cells. However, in the case of Tasmanian devils, their immune system is unable to detect these cancer cells as foreign due to the limited genetic diversity within the devil population. The contagious cancers, specifically DFT1 and DFT2, exploit this vulnerability and are transmitted through biting, a common behavior among devils, leading to rapid spread among the population.
\\
\hline
Q3 &  \textcolor{blue}{[Quesion]}:  I appreciate your clarification on the contagious nature of these cancers in Tasmanian devils. For my third question, how have these cancers (DFT1 and DFT2) evolved over time, especifically in terms of increased transmissibility or effectiveness? \\
\hline

A3 & \textcolor{blue} {[Answer]}: I'm glad my answers have been helpful. Regarding the evolution of DFT1 and DFT2, you can envision it as similar to software updates, enhancing their functionality over time. When DFT1 initially emerged, it went through a "superspreader" event that led to six major DFT1 variants. These adaptations allowed the cancer cells to have improved transmission and survival capabilities. Meanwhile, DFT2, though genetically similar to DFT1, has a substantially faster mutation rate. Its rapid evolution could be likened to an accelerated, advanced software system refining at an expedited pace. This increased mutation capability may provide DFT2 a competitive edge and allow it to spread more effectively. However, researchers are yet to determine if these mutations convey a specific selective advantage.\\
\hline
Q4 &  \textcolor{blue}{[Quesion]}: Thank you for further detailing the evolution of DFT1 and DFT2. My fourth question is, what possible applications or considerations might there be for creating a medical intervention for these cancers in Tasmanian devils, such as vaccines or treatments based on these recent findings?\\
\hline

Q4 & \textcolor{blue} {[Answer]}:  Certainly. Drawing from the artificial intelligence domain, the ultimate goal is often optimization or finding the best solution for a problem. In this context, these recent findings about DFT1 and DFT2 help us understand the genetic vulnerabilities, origin, and evolutionary patterns of these cancers. This knowledge offers a robust dataset that can influence the development of informed, targeted medical interventions for Tasmanian devils.
 
 Scientists can use this information to devise specific strategies, such as engineering vaccines or treatments that counteract the fast mutation rate of these cancers or appropriately modify the peculiar immune system loopholes in devils. The key is leveraging the insights gleaned from genomic analysis in a manner similar to developing custom plugins or patches in AI to address the anomalies driving DFT1 and DFT2 tumor growth and transmission more effectively.
\\
\hline
Q5 &  \textcolor{blue}{[Quesion]}: As someone working in the field of artificial intelligence, my fifth question is connected to my area of expertise. Can large-scale genomic sequence data and machine-learning algorithms be employed together to make better predictions or advance our understanding of these transmissible cancers in Tasmanian devils and potentially reveal new treatment methods?
\\
\hline

A5 & \textcolor{blue} {[Answer]}: Absolutely! Large-scale genomic sequence data and machine learning algorithms can have synergistic effects in analyzing and understanding these transmissible cancers. To put it in context, consider genomic sequences as immense databases and machine learning algorithms as highly specialized data mining tools.
 
 Machine learning algorithms can be trained to recognize patterns, discover correlations, and possibly identify novel tumor characteristics from the genomic data, similar to detecting useful features within vast datasets in the AI realm. Furthermore, these algorithms may help uncover hidden links between genes or genetic mutations and cancer progression, improving our understanding of these diseases while potentially revealing novel targets for therapeutics or helping create personalized medical interventions.
 
 In essence, the combination of genomic data and machine learning algorithms forms a powerful resource to advance the investigation of transmissible cancers in Tasmanian devils and guide treatment development.
\\
\hline
Q6 &  \textcolor{blue}{[Quesion]}: Thank you for demonstrating the potential benefits of utilizing large-scale genomic sequence data and machine learning algorithms in the study of transmissible cancers. For my second-to-last question, how might I use the intricate modeling techniques in AI or machine learning to better grasp the possible futures of Tasmanian devil populations or contain DFT1 and DFT2 by predicting their spread and evolution?\\
\hline

A6 & \textcolor{blue} {[Answer]}: In the context of AI and machine learning, you can leverage intricate modeling techniques to better understand the potential futures of Tasmanian devil populations and address the spread of DFT1 and DFT2. Utilize the empirical genomic data on these cancers, their mutation and transmission patterns, and variables such as devil populations, geographic data, and environmental factors as inputs to construct sophisticated predictive models.
 
 These models would help forecast how DFT1 and DFT2 could evolve, spread, and interact with the host population, similar to running complex simulations in AI using diverse datasets to analyze plausible outcomes. Actively adapting machine learning algorithms could identify high-risk areas, transmission hot-spots, or anticipate population vulnerabilities.
 
 Armed with such predictive insights, conservationists, and domain experts can undertake pre-emptive measures, design targeted interventions, and efficiently optimize strategies to prevent the aggressive spread of these cancers and ultimately conserve and protect Tasmanian devils from possible population collapse.\\
\hline
Q7 &  \textcolor{blue}{[Quesion]}: Thank you for demonstrating the potential benefits of utilizing large-scale genomic sequence data and machine learning algorithms in the study of transmissible cancers. For my second-to-last question, how might I use the intricate modeling techniques in AI or machine learning to better grasp the possible futures of Tasmanian devil populations or contain DFT1 and DFT2 by predicting their spread and evolution?\\
\hline

A7 & \textcolor{blue} {[Answer]}: In the context of AI and machine learning, you can leverage intricate modeling techniques to better understand the potential futures of Tasmanian devil populations and address the spread of DFT1 and DFT2. Utilize the empirical genomic data on these cancers, their mutation and transmission patterns, and variables such as devil populations, geographic data, and environmental factors as inputs to construct sophisticated predictive models.
 
 These models would help forecast how DFT1 and DFT2 could evolve, spread, and interact with the host population, similar to running complex simulations in AI using diverse datasets to analyze plausible outcomes. Actively adapting machine learning algorithms could identify high-risk areas, transmission hot-spots, or anticipate population vulnerabilities.
 
 Armed with such predictive insights, conservationists, and domain experts can undertake pre-emptive measures, design targeted interventions, and efficiently optimize strategies to prevent the aggressive spread of these cancers and ultimately conserve and protect Tasmanian devils from possible population collapse.\\
\hline
\end{longtblr}

\begin{longtblr}[
  caption = {CONA Analogy RSP for QA1 on presentation 'Tasmanian Devil Cancer \cite{conroy_2023}' with audience AI Expert},
  label = {Touch_Art_Text},
]{
  colspec = {|X[0.7]|X[12]|},
  rowhead = 1,
  hlines,
  row{even} = {gray9},
  row{1} = {yellow},} 
Part & RSP Text\\

A1 &  \textcolor{blue}{[Answer]}:  As an Artificial Intelligence Expert, the terms DFT1 and DFT2 lie outside my expertise. However, based on the information you presented, I will attempt to create an analogy to the best of my ability.
 
 Consider two strains of a computer virus, Virus1 and Virus2. Despite originating separately, they are similar in purpose - to spread and infect more systems. Virus1 spreads throughout most of an interconnected network and has multiple variations due to years of existence. On the other hand, Virus2 begins later and is primarily confined to one part of the network, but mutates much faster. Understanding the origins, development, and possible future shifts in these viruses can help us mitigate their consequences and protect our computer networks better.
 
 This is my first version of the analogy. Please provide your feedback, and I will make adjustments accordingly.\\
\hline

S1 & \textcolor{blue} {[Suggestion]}: 1. Maintain relatability: Choose an understandable scenario, like a computer virus or a disease.
 2. Emphasize separate emergence: Mention that the analogy objects have different origins, similar to DFT1 and DFT2.
 3. Include impact: Discuss how both analogy objects can cause significant harm to the system, as the devil facial tumors do to the Tasmanian devil populations.
 4. Describe spread: Explain that the analogy objects can spread and mutate through contact or transmission, as the contagious facial cancers do in real life.
 5. Consider the evolutionary aspect: Address the evolutionary aspect of the devil facial tumors by including a corresponding element in the analogy.
 
 Keep those suggestions in mind and adjust your analogy to effectively convey the natural process seen in Tasmanian devils with their transmissible facial cancers.\\
\hline
A2 &  \textcolor{blue}{[Answer]}:  Consider two strains of malicious software (malware), Malware1 and Malware2, which affect computer networks. Malware1 emerged quite some time ago and has since spread through a massive network, generating multiple variants along the way due to its many transmissions. Conversely, Malware2 arose more recently and primarily invades a smaller section of the network, but mutates at an accelerated rate. Similar to the Tasmanian devil facial tumors, the two malware strains emerged separately, causing significant disturbances to their hosts, and evolved over time through spreading and mutating via transmission. By studying the origins, evolutionary paths, and propagation trends of Malware1 and Malware2, experts can better decipher potential future dangers and secure computer networks against such attacks.

This is my modified version of the analogy, which integrates the provided suggestions. Please let me know if there are additional changes to be made.\\
\hline

S2 & \textcolor{blue} {[Suggestion]}:  1. Clarify the similarities and differences between Malware1 and Malware2, parallel to the characteristics of DFT1 and DFT2.
 2. Emphasize the impact of the malware strains' emergence and evolution on their host networks, analogous to the Tasmanian devil population.
 3. Explain the significance of understanding the mechanisms behind malware spread and mutation for devising protective measures, akin to conserving and treating the Tasmanian devil population.
 4. Highlight the importance of research to predict future consequences and trends for both the computer networks and Tasmanian devil populations.\\
\hline
A3 &  \textcolor{blue}{[Final Answer]}:  Imagine two strains of malware impacting computer networks, Malware1 and Malware2, each with separate origins and distinct characteristics. Malware1 emerged earlier and rapidly spread across a vast network, birthing numerous variants. Malware2 surfaced more recently and focused on a smaller network section, marked by its accelerated mutation rate. The emergence and evolution of these malware strains negatively affect the host networks, just as DFT1 and DFT2 have inflicted the Tasmanian devil population. Studying both the malicious origin and their evolutionary paths can aid in understanding mechanisms underpinning their spread and mutations, ultimately contributing to developing protective strategies for conserving and treating host networks and the Tasmanian devil population. Research is vital for predicting long-term implications and trends in terms of system security and Tasmanian devil populations.\\
\hline
\end{longtblr}

\begin{longtblr}[
  caption = {CONA ProbSolving RSP for QA5 on presentation 'Tasmanian Devil Cancer \cite{conroy_2023}' with audience AI Expert},
  label = {Touch_Art_Text},
]{
  colspec = {|X[0.7]|X[12]|},
  rowhead = 1,
  hlines,
  row{even} = {gray9},
  row{1} = {olive9},} 
Part & RSP Text\\

A1 &  \textcolor{blue}{[Answer]}:  As an artificial intelligence expert, I'm afraid I don't have the in-depth understanding and knowledge required to answer questions about transmissible cancers in Tasmanian devils. My expertise is in AI and related technologies. If you have any questions about artificial intelligence, I'd be happy to help.\\
\hline

S1 & \textcolor{blue} {[Suggestion]}: 1. Reframe the context of the question to focus on genomic data analysis and machine learning's potential contributions to cancer research.
 2. Explore current applications of AI and machine learning in cancer research, and discuss similarities or differences with transmissible cancers.
 3. Delve into the potential challenges of employing large-scale genomic sequence data and machine learning algorithms in the context of Tasmanian devils' transmissible cancers.
 4. Discuss areas of artificial intelligence and machine learning research that could potentially be applied to such problems in the future.
 5. Highlight examples of successful interdisciplinary collaboration between AI experts and cancer biologists or geneticists, emphasizing their impact on devising treatment methods or preventive strategies.\\
\hline
A2 &  \textcolor{blue}{[Answer]}:  CWhile my expertise is in artificial intelligence rather than cancer research, I can offer some suggestions on how large-scale genomic sequence data and machine-learning algorithms could contribute to the understanding and potential treatment methods for transmissible cancers in Tasmanian devils.
 
 First, genomic data analysis and machine learning can be useful in identifying patterns and establishing connections between genetic factors and the emergence or spread of the cancer. This information can help inform strategies for prevention or containment. In a more general sense, AI and machine learning are already in use in cancer research by identifying genetic mutations responsible for the disease, prognosis prediction, analyzing treatment effects, etc. These applications could potentially extend to transmissible cancers in Tasmanian devils.
 
 Challenges regarding the use of AI and machine learning for these specific transmissible cancers may include the rarity of the disease and limited resources for study. However, interdisciplinary collaboration between AI experts, cancer biologists, and geneticists has been valuable in addressing other complex scientific problems, leading to innovative treatment methods and prevention strategies.
 
 In the future, AI research areas, such as reinforcement learning, deep learning, or natural language processing, might be used to better understand and tackle Tasmanian devil transmissible cancers. However, expertise from biologists, geneticists, and experts specifically studying these cancers would be crucial for assessing the applicability of artificial intelligence techniques in this context.\\
\hline

S2 & \textcolor{blue} {[Suggestion]}:  1. Focus on simplifying the technical jargon to make the answer more accessible to the public. For example, provide or define terms like "genomic sequence data, reinforcement learning, and natural language processing" in an easy-to-understand manner.
 
 2. Establish a clear connection between using AI technology and transmissible devil facial cancers; make sure the application significant implications become explicit.
 
 3. Provide examples or case studies relevant to the AI algorithms and methods applied to similar challenges in cancer research or analyze the treatment effect. This would broaden the understanding of practical situations.
 
 4. Stress on the importance of interdisciplinary collaboration: how cancer researchers, biologists, geneticists can work together cooperation with AI experts for mutual understanding and knowledge gain.
 
 By making these adjustments, the answer will become more coherent, relatable, and engaging for the audience. Ultimately, readers should be able to understand the applications of AI and its potential for providing new insights into devil facial transmissible cancers.\\
\hline
A3 &  \textcolor{blue}{[Answer]}:  Large-scale genomic sequence data refers to large volumes of detailed genetic information collected from different sources. Machine learning algorithms, on the other hand, are AI techniques that can learn from data and make predictions based on patterns they detect. Both of these tools can potentially help advance our understanding and potential treatment methods for transmissible cancers in Tasmanian devils.
 
 An example of a relevant AI algorithm application in cancer research is deep learning, which excels at finding subtle patterns in complex data. By analyzing genomic data to detect genetic factors linked to the emergence or progression of these cancers, researchers can contribute to devising potential treatment strategies.
 
 Reinforcement learning, a method where AI learns the best action from continuous feedback, might also play a role in simulating the transmission dynamics of devil facial cancers and suggest effective prevention measures.
 
 Natural language processing, which helps AI understand human language, can assist in processing research papers, identifying common findings, and highlighting areas of interest to guide researchers to better treatment methods and management of the cancer.
 
 Collaboration is essential between AI experts, cancer researchers, biologists, and geneticists working together to understand devil facial transmissible cancers. Existing knowledge of AI potentials in the context of cancer research can be fruitfully applied, leading to innovative treatment and prevention strategies for devil facial tumors. These interdisciplinary efforts would better pave the way for AI technology in providing significant insights into tackling devil facial transmissible cancers.\\
\hline

S3 & \textcolor{blue} {[Suggestion]}:  1. Provide a brief overview of devil facial tumours (DFT1 and DFT2) to set the context before discussing genomic sequence data and machine learning.
 2. Describe the basics of genomic sequence data and machine learning algorithms, ensuring the concepts are easily comprehensible for non-expert audiences.
 3. Focus on specific applications of machine learning algorithms in genomic data analysis that directly address analyzing and predicting transmissible cancers in the Tasmanian devils context.
 4. Offer examples of impactful interdisciplinary efforts between AI experts and cancer researchers that have led to valuable insights in understanding, predicting, or treating cancers.
 5. Address potential challenges and limitations associated with implementing genomic sequencing data and AI algorithms in the context of Tasmanian devil tumours, while providing a balanced perspective on expectations and tangible outcomes.
 6. Highlight the long-term goals and broader implicationsâ€”positive or negativeâ€”of incorporating AI technologies in the field of cancer research, with a focus on Tasmanian devil transmissible cancers.\\
\hline
A4 &  \textcolor{blue}{[Final Answer]}:
Contagious facial cancers affecting Tasmanian devils, known as devil facial tumour 1 (DFT1) and devil facial tumour 2 (DFT2), have led to a significant decrease in their population. Genomic sequence data comprises large volumes of detailed genetic information, while machine learning algorithms involve AI techniques that can learn from data to make predictions or recognize patterns.
 
 Applying machine learning algorithms on genomic sequence data can potentially help advance understanding and identifying potential treatments for transmissible cancers in Tasmanian devils. Algorithms could analyze patterns in genomic data and devise strategies to tackle cancers.
 
 For example, AI techniques, such as deep learningâ€”which finds patterns in complex datasetsâ€”could detect genetic factors specifically linked to the emergence or progression of these transmissible cancers. Real-life interdisciplinary efforts between AI experts and cancer researchers have yielded valuable insights into understanding, predicting, and treating cancers.
 
 However, there could be challenges in analyzing large volumes of genomic data or training AI algorithms on relatively uncommon diseases like DFT1 and DFT2. It may be essential to scale expectations when working on this particular scenario.
 
 Incorporating AI technologies for long-term goals could broaden the implications of AI in the fieldâ€”both positive and negative. Addressing AI opportunities, challenges, and limitations in the context of Tasmanian devil facial cancers is necessary to provide realistic outcomes and moving towards innovative strategies in cancer research.\\
\hline
\end{longtblr}

\newpage

\begin{longtblr}[
  caption = {CONA Dilemma RSP for QA7 on presentation 'Tasmanian Devil Cancer \cite{conroy_2023}' with audience AI Expert},
  label = {Touch_Art_Text},
]{
  colspec = {|X[0.7]|X[12]|},
  rowhead = 1,
  hlines,
  row{even} = {gray9},
  row{1} = {yellow},} 
Part & RSP Text\\

A1 &  \textcolor{blue}{[Answer]}:  Based on the presented information, our position as proponents for increased conservation efforts emphasizes the alarming decline in Tasmanian devil populations due to contagious facial cancers, DFT1 and DFT2. Intensifying conservation initiatives can help mitigate the impact of these cancers by actively monitoring the populations, managing the disease, and implementing protective measures such as captive breeding programs. These actions can safeguard the genetic diversity of this endangered species while aiding in the preservation of their natural habitat. Additionally, research into the genomics of these diseases can bolster our understanding and facilitate the development of possible treatments, ultimately leading to better-informed conservation decisions. Collectively, these steps will ensure the survival and future resilience of the Tasmanian devil population while combatting the devastating effects of facial cancers.\\
\hline

S1 & \textcolor{blue} {[Suggestion]}:  1. Acknowledge the importance of natural evolutionary processes and aim for minimal interference in conservation efforts.
 2. Employ cautious monitoring and gentle disease management techniques rather than invasive measures.
 3. Establish clear objectives for captive breeding programs to maintain genetic diversity and ensure only healthy individuals are reintroduced to the wild.
 4. Emphasize long-term sustainability in conservation strategies rather than immediate interventions.
 5. Foster collaboration between stakeholders, researchers, and wildlife experts to enable adaptive management of the conservation efforts.\\
\hline
A2 &  \textcolor{blue}{[Answer]}: In light of the suggestions, our modified viewpoint acknowledges the importance of natural evolutionary processes and accentuates minimal intervention in Tasmanian devil conservation efforts. We understand that intensified conservation initiatives should include cautious monitoring, gentle disease management techniques, and clear objectives for captive breeding programs to maintain genetic diversity. Our advocacy for increased conservation focuses on long-term strategies and sustainability. Furthermore, we encourage collaboration among stakeholders, researchers, and wildlife experts to promote informed decision-making, adaptive management, and combined efforts to protect the Tasmanian devil population while also addressing the challenges posed by contagious facial cancers.\\
\hline

S2 & \textcolor{blue} {[Suggestion]}:   1. Implement cautious disease containment and mitigation strategies, preserving the natural evolutionary processes for Tasmanian devils.
 2. Reevaluate the balance between field efforts and captive breeding programs, considering their impacts on genetic diversity.
 3. Emphasize and encourage local engagement, education, and awareness campaigns to facilitate positive human-devil interactions.
 4. Advocate for further research on the infectious facial cancers to better understand their etiology and enable targeted interventions.
 5. Establish periodical evaluations to assess conservation strategies and adapt if any unforeseen negative consequences emerge.\\
\hline
A3 &  \textcolor{blue}{[Answer]}: Taking into account the recent suggestions, our updated viewpoint underscores the need to carry out cautious containment methods and maintain natural evolutionary processes. When assessing conservation strategies, we plan to maintain a healthy balance between field efforts and controlled captive breeding programs. Furthermore, we believe in the value of amplifying local engagement, educational initiatives, and enhancing public awareness to foster positive human-devil coexistence.
 
 As we continue advocating for further research related to the infectious facial cancers, we intend to refine our understanding of these deadly diseases and support better-targeted interventions where needed. Lastly, to ensure the success of conservation approaches, we emphasize the importance of periodic evaluations, adjusting plans to mitigate any unforeseen detrimental consequences. This ensures that we can both protect Tasmanian devils while acknowledging the vital balance between human intervention and natural processes.\\
\hline

S3 & \textcolor{blue} {[Suggestion]}:   1. Acknowledge that natural evolutionary processes have a role in shaping Tasmanian devil populations, but stress the urgency of action considering the rapid spread of facial cancers.
 2. Encourage monitored conservation interventions without overstepping the bounds of natural ecological dynamics. 
 3. Incorporate local stakeholder perspectives to establish sustainable strategies for the conservation of Tasmanian devils in the wild.
 4. Collaborate on the development of contingency plans to manage unforeseen issues arising from implemented conservation measures.
 5. Emphasize data-driven decision-making and recalibration of strategies as new information becomes available.\\
\hline
A3 &  \textcolor{blue}{[Final Answer]}: In response to the suggestions, our revised viewpoint acknowledges the significance of natural evolutionary processes in shaping the Tasmanian devil population. However, given the rapid spread of contagious facial cancers, we recognize the urgency of action. We encourage monitored conservation interventions, ensuring they do not disproportionally disrupt ecological dynamics.
 
 For the sake of sustainable conservation, it is essential to engage with and incorporate local stakeholder viewpoints in discussions and strategies affecting Tasmanian devils in wild populations. As a measure of prudence, we advocate for the development of contingency plans managing potential unforeseen consequences of employed conservation actions.
 
 Finally, we maintain our focus on employing a data-driven approach, consistently adopting decision-making practices based on current scientific discoveries and tweaking strategies as necessary when new findings arise, fostering a more comprehensive and structurally neutral viewpoint on this crucial issue.
\\
\hline
\end{longtblr}

\end{document}